\documentclass[journal]{IEEEtran}

\usepackage{cite}
\usepackage[hidelinks]{hyperref}
\usepackage{balance}

\usepackage{orcidlink}
\ifCLASSINFOpdf
\else
   \usepackage[dvips]{graphicx}
\fi
\usepackage{url}

\usepackage{graphicx}
\usepackage{algorithm}
\usepackage{algorithmic}
\usepackage{amsmath}

\begin{document}

\title{Resource-Efficient Motion Control for Video Generation via Dynamic Mask Guidance}

\author{Sicong Feng\orcidlink{0009-0008-8912-6887}, Jielong Yang\orcidlink{0000-0001-5853-6316}, and Li Peng\orcidlink{0000-0001-8085-9840}

\thanks{Sicong Feng, Jielong Yang and Li Peng are with the Jiangnan University, Wuxi 214122, China (e-mail: 6231913014@stu.jiangnan.edu.cn; jyang022@e.ntu.edu.sg; pengli@jiangnan.edu.cn).}
}

\maketitle

\begin{abstract}
Recent advances in diffusion models bring new vitality to visual content creation. However, current text-to-video generation models still face significant challenges such as high training costs, substantial data requirements, and difficulties in maintaining consistency between given text and motion of the foreground object. To address these challenges, we propose mask-guided video generation, which can control video generation through mask motion sequences, while requiring limited training data. Our model enhances existing architectures by incorporating foreground masks for precise text-position matching and motion trajectory control. Through mask motion sequences, we guide the video generation process to maintain consistent foreground objects throughout the sequence. Additionally, through a first-frame sharing strategy and autoregressive extension approach, we achieve more stable and longer video generation. Extensive qualitative and quantitative experiments demonstrate that this approach excels in various video generation tasks, such as video editing and generating artistic videos, outperforming previous methods in terms of consistency and quality. Our generated results can be viewed in the supplementary materials.
\end{abstract}

\begin{IEEEkeywords}
Motion pattern control, latent diffusion models, video generation.
\end{IEEEkeywords}

\IEEEpeerreviewmaketitle

\section{Introduction}

\IEEEPARstart{R}{ecent} years witness the maturation of image generation technology and its widespread application across numerous domains, including style transfer \cite{cheng2021one},\cite{deng2018enhancing},\cite{zhou2022neural} and text-to-image generation. Notably, diffusion-based methods \cite{ho2020denoising},\cite{song2020denoising},\cite{song2020score} achieve remarkable breakthroughs in text-prompted image generation. Models such as DALLE2 \cite{ramesh2022hierarchical}, Stable Diffusion \cite{rombach2022high}, and Imagen \cite{saharia2022photorealistic} demonstrate the ability to generate diverse and high-quality images guided by text prompts. Given the success of text-to-image generation, researchers turn their attention to text-to-video generation (T2V). However, unlike image generation, video generation presents greater challenges as it requires handling both static features and motion dynamics simultaneously.

\begin{figure}[!t]
    \includegraphics[width=\columnwidth]{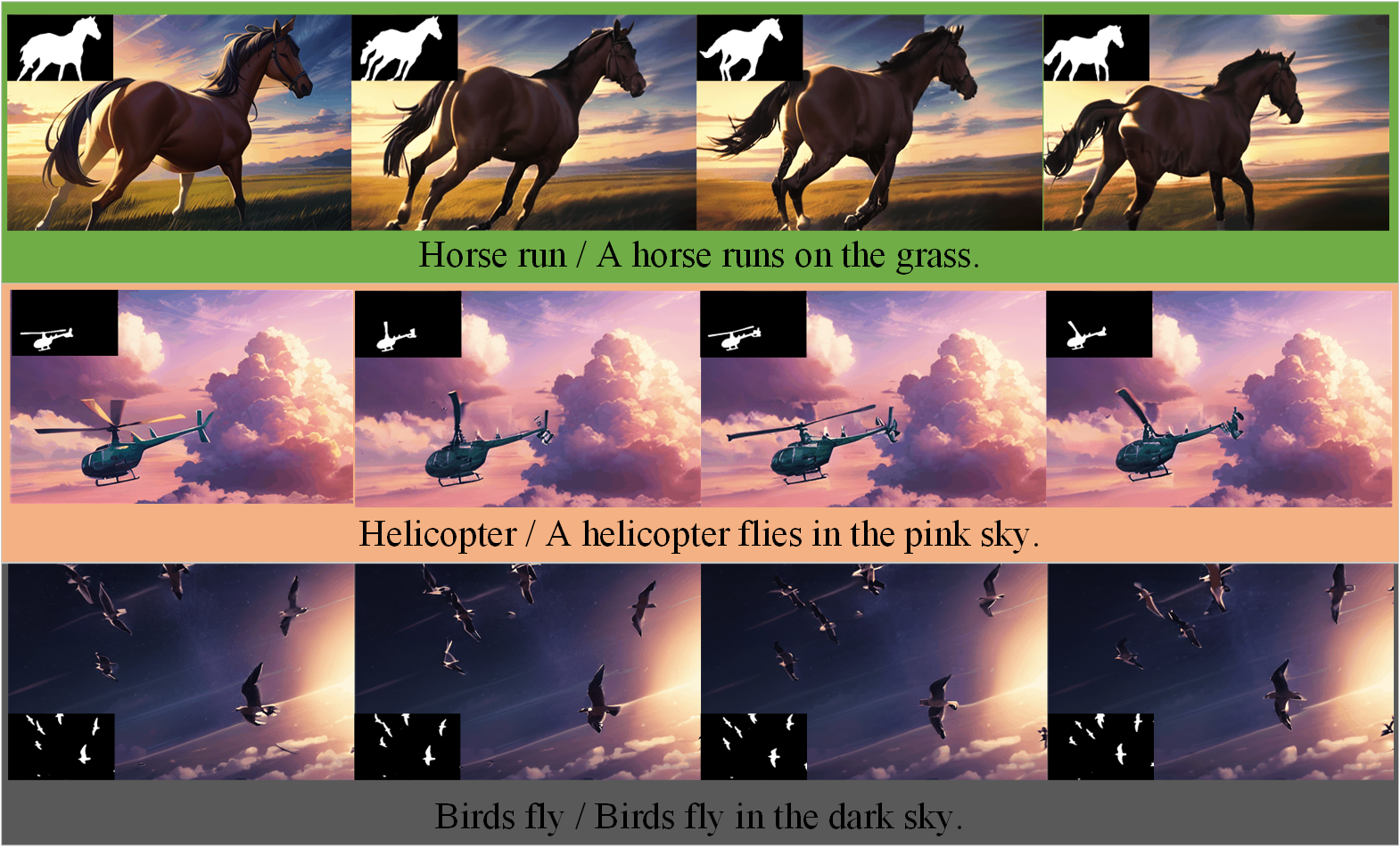}
    \caption{Our model generates various videos consistent with the foreground mask and text prompts, delivering satisfactory results.}
    \label{fig1}
\end{figure}

\begin{figure}[!t]
    \includegraphics[width=\columnwidth]{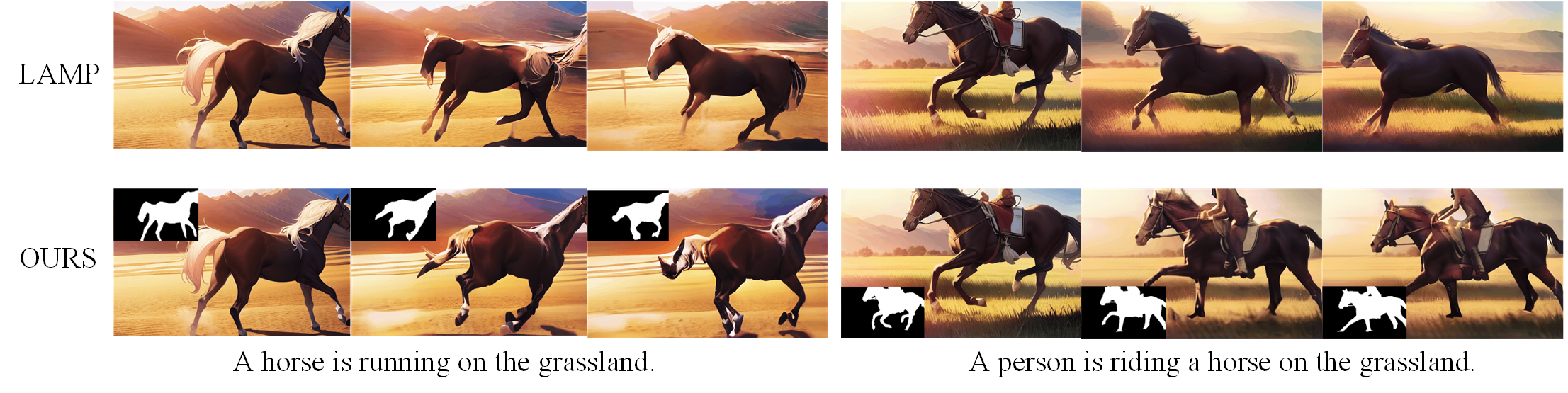}  
    \caption{In LAMP\cite{wu2024lamp}, the model fails to capture the horse's movement direction and distinguish foreground-background properly. Our model addresses these issues by accurately tracking motion and maintaining clear foreground-background separation.}
\label{fig2}
\end{figure}

Several existing approaches utilize extensive video training \cite{blattmann2023align},\cite{esser2023structure},\cite{ho2022imagen}, but these approaches demand substantial computational resources and time, which limits their scalability. To address this, approaches like Tune-A-Video \cite{wu2023tune} reduce training overhead through single-video fine-tuning, yet suffer from low efficiency and overfitting. Building on this, LAMP \cite{wu2024lamp} propose a few-shot training method that effectively improves video quality and freedom, but still faces two major challenges(see Fig. \ref{fig2}): (i) Imprecise foreground positioning and motion capture, leading to subsequent errors in interpreting foreground object movement; (ii) Unnatural fusion between moving foreground and background, preventing the model from distinguishing between text-described subjects and backgrounds, resulting in foreground object disappearance.

\begin{figure*}[!t]
    \includegraphics[width=\textwidth]{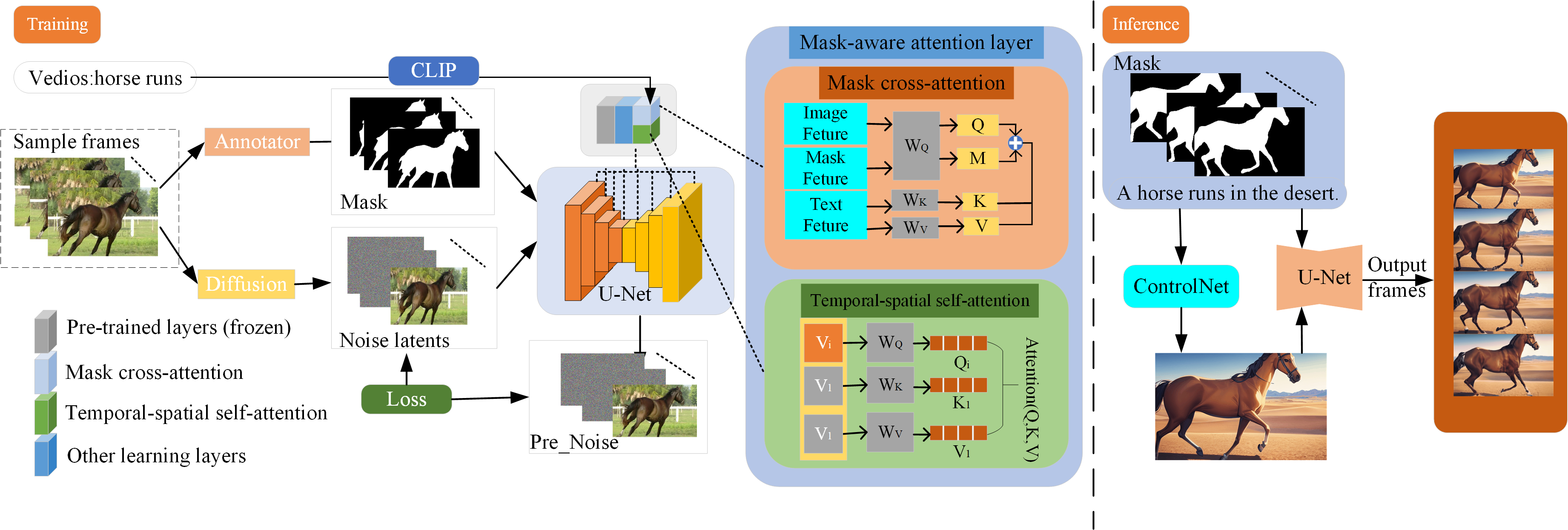}  
    \caption{Overall framework of our mask-guided video generation method. We apply trainable temporal-spatial self-attention and mask cross-attention within the U-Net, enabling the model to focus more on the foreground. 
}
\label{fig3}
\end{figure*}

To address these issues, introducing control signals presents a potential solution. We propose a novel mask-guided video generation method. Unlike previous works \cite{wu2023tune}, \cite{wu2024lamp}, we introduce a foreground mask branch in the network to adjust attention for foreground positioning and motion capture. As shown in Fig. \ref{fig2}, our model effectively captures the foreground position and motion trajectory defined by the text. During inference, we utilize the provided mask sequence to guide motion generation, which effectively maintains the distinction between foreground and background elements while preventing foreground objects from unexpectedly disappearing. Following this approach, we can generate longer video through autoregression. Additionally, we utilize the first frame's latent representation as shared noise to improve video generation quality and stability. 

In summary, our contributions are as follows: \textbf{(i)} This text-to-video method can be trained in a single GPU using only a small video dataset. \textbf{(ii)} Our model introduces an innovative mask-aware attention layer during training, which allows for more precise capture of the foreground region. During generation, by providing a mask sequence to control the video generation, it effectively prevents the blending of foreground and background, ensuring clear separation and stable presentation of the foreground in the video. \textbf{(iii)} We empirically demonstrate that our model achieves excellent results in both consistency and quality. As a byproduct of using the first-frame conditional generation strategy, our method is capable of generating long videos\footnote{ Same as the definition in Control-A-
Video\cite{chen2023control} } while ensuring the continuity and consistency of the target object's motion.

\section{RELATED WORK}

Text-to-image generation makes significant breakthroughs. Methods like DenseDiffusion \cite{kim2023dense} further address complex text generation requirements, while ControlNet \cite{zhang2023adding} introduces additional controllable conditions to Stable Diffusion, such as depth maps, poses, and edges. In our work, we incorporate masks during first frame generation using ControlNet \cite{zhang2023adding} to constrain text-corresponding positions, providing content guidance for subsequent video generation.

As image generation technology matures, researchers gradually extend their focus to video generation, achieving notable results through models such as Make-A-Video \cite{singer2022make}, MagicVideo \cite{zhou2022magicvideo}, VideoComposer \cite{wang2024videocomposer}, CogVideo \cite{hong2022cogvideo}, and AnimateDiff \cite{guo2023animatediff}, but these approaches typically require substantial computational resources for training. 
In contrast, zero-shot methods such as Text2Video-Zero \cite{khachatryan2023text2video}, ControlVideo \cite{zhang2024controlvideo}, and Style-A-Video \cite{huang2024style} eliminate the need for fine-tuning. However, their reliance on large-scale pre-trained models and general-purpose data limits their capability for fine-grained generation in specific domains. Control-A-Video \cite{chen2023control} and MoonShot \cite{zhang2024moonshot} draw inspiration from ControlNet \cite{zhang2023adding}, guiding video generation through depth maps and edge information to improve quality, but still consume considerable resources. To address these limitations, we extend LAMP \cite{wu2024lamp} by introducing foreground mask motion sequences for video generation guidance. Our approach requires only few-shot samples and single GPU training, while achieving enhanced generation diversity through varied initial frames.

\section{Method}
\label{sec:guidelines}


\subsection{Mask-Guided Training}

Latent Diffusion Models (LDMs) \cite{rombach2022high} are an efficient variant of diffusion models that operate in latent space. Specifically, the encoder $E$ is used to extract a latent feature representation $x_0=E(I)$ from an image $I$. During the forward process, noise is gradually added to the latent features, which eventually approach a Gaussian distribution as time step $t$ increases. In the reverse denoising process, a U-Net is trained to predict the added noise using DDIM \cite{song2020denoising} and conditional inputs (such as text prompts $c_p$). The training objective is to minimize the difference between predicted and actual noise, with the loss function expressed as:

\begin{equation}  
	L_{simple}=E_{x_0,\varepsilon,t}\left[\left\|\varepsilon-\varepsilon_\theta(x_t,t,c_p)\right\|_2^2\right],
\end{equation}
where $\varepsilon$ is the noise from a standard normal distribution, and $\varepsilon_{\theta}$ is the noise predicted by the neural network.

Based on our observations, the first frame of a video often contains key information of the entire video. Therefore, we use the content of the first frame as a condition to generate the subsequent frames. As shown in Fig \ref{fig3}, during training,  we represent the sampled $n$ video frames as $V=\{f_i|i=1,...,n\}$. These frames are embedded into latent space using an encoder and we obtain $X=\{x_i|i=1,...,n\}$. We then keep the first frame $x_1$ unchanged and apply a forward diffusion process to the subsequent frames ${x_2,…,x_n }$ to obtain the noisy video frame sequence ${\varepsilon_2,\dots,\varepsilon_n}$. The loss function can be expressed as:
\begin{equation}  
	L=E_{x,\varepsilon\sim N(0,I),t,c_p}\left[\left\|\varepsilon_{2:n}-\varepsilon_{2:n}^\theta(x_t,t,c_p)\right\|_2^2\right],
\end{equation}
where $\varepsilon_{2:n}$ represents the noisy video frame sequence from the 2nd to the $n$-th frame and $\varepsilon_{2:n}^\theta$ is the noise predicted by the neural network at time step $t$ conditioned on the input $c_p$ (e.g., text prompts). Using this method, the model gains the capability to generate a video with the motion pattern of the video set according to the first frame.

During inference, we input the first frame $m_1$ of the mask motion sequence, the text prompt $c_p$, and random Gaussian noise $x_1$ into the ControlNet \cite{zhang2023adding} model to obtain the initial frame $v_1$:
\begin{equation}  
	v_1=\text{ControlNet}{\left(x_1,c_p,m_1\right)},
\end{equation}

After obtaining the desired first frame, we encode the first frame and obtain $E(v_1 )$. The subsequent motion sequence frames are generated using:
\begin{equation}  
	v=\text{MaskVideo}\left(x,c_p,m,E(v_1)\right),
\end{equation}

\begin{figure}[!t]

    \includegraphics[width=\columnwidth]{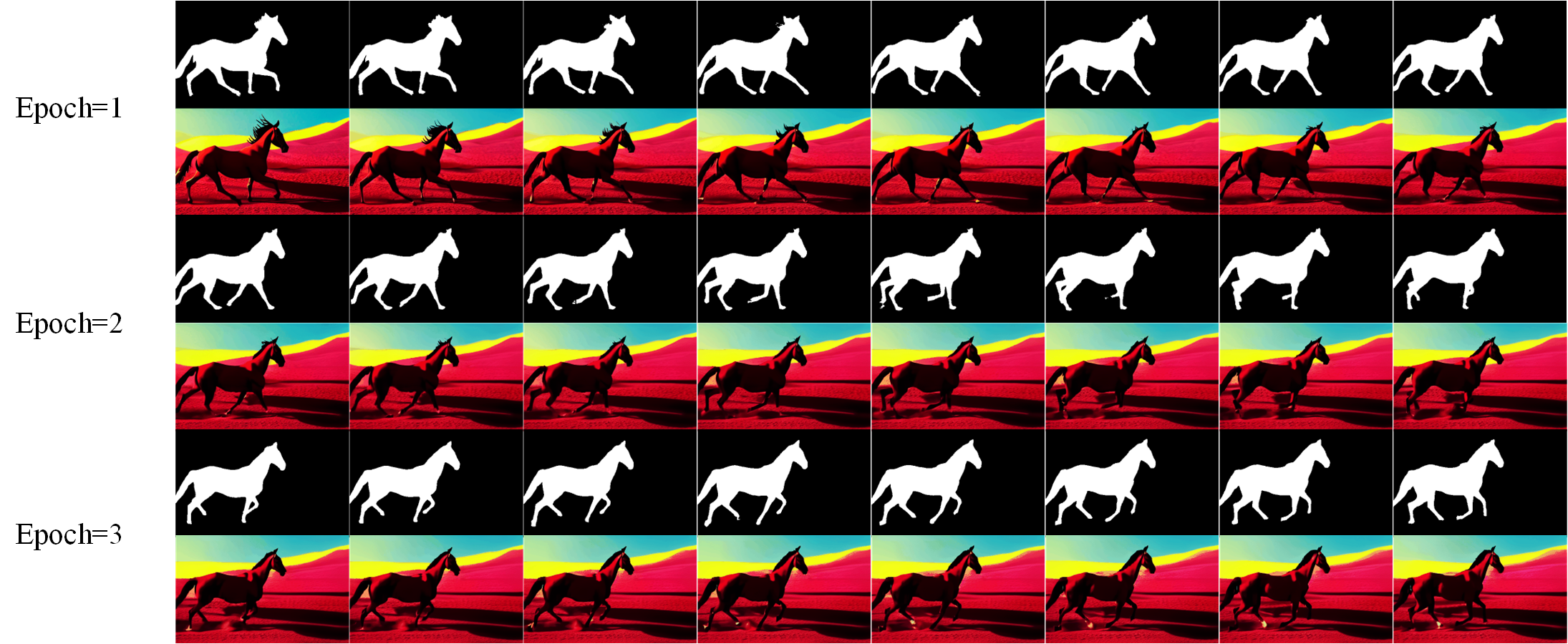}  
    \caption{\textbf{Auto-Regressive Generation.} Our model is capable of generating long videos. Given the text prompt "A horse runs across a flat desert plain under a midday sun in a pop art painting style," the video frames are generated using a first-frame-based method, producing a 24-frame video after three epochs.}
\label{fig5}
\end{figure}

The video generation method we develop, based on the first-frame condition, enables precise control over the content of dynamic videos and can produce long videos using an autoregressive approach, as shown in Fig \ref{fig5}.

\subsection{Adapt T2I Models to Video}

\textbf{Mask-aware Attention Layer.} To integrate with the proposed pipeline and facilitate subsequent frames referencing the conditions established by the first frame, we introduce a new temporal-spatial self-attention. To ensure consistency, all key and value features are derived from the first frame, which can be written as:
\begin{equation}  
	\text{SelfAttn}(Q^i,K^1,V^1)=\text{Softmax}\left(\frac{Q^i(K^1)^T}{\sqrt{d}}\right)V^1,
\end{equation}
where $i=\in{1,\dots,n}$ indicates that the extracted feature map comes from the $i-th$ frame, and $Q,K,V\in R^{BNS\times HW\times\frac CS}$, with $B$ representing the batch size, $N$ the number of frames, $H$ the height, $W$ the width, $C$ the number of channels, and $S$ the number of attention heads. The attention scores establish the connection between the first frame and subsequent frames.

Additionally, to better match the foreground with the text, we propose mask cross-attention, which incorporates the mask into the calculation of the attention scores in cross-attention. Specifically, we first project the the original feature map $v$ from the U-Net, the foreground mask $m$, and the text embedding vector $c$, using the following equations:
\begin{equation}  
	Q=w^Q\cdot v,\quad K=w^K\cdot c,\quad V=w^V\cdot c,\quad M=w^Q\cdot m,
\end{equation}
where $M\in R^{BNS\times HW\times\frac CS}$, and the cross attention can be written as:
\begin{equation}  
	\text{CrossAttn}{\left(Q,K,V,M\right)}=\text{Softmax}\left(\frac{QK^T+MK^T}{\sqrt{d}}\right)V,
\end{equation}

With this method, the degree of matching between text features and the motion of the foreground object is significantly improved, allowing the model to more accurately identify and capture the motion of the foreground object, achieving high consistency between text features and the motion of the foreground object. 


\textbf{First-frame shared sampling strategy.} During inference, we find that if the shared noise is a randomly sampled $\varepsilon_s{\sim}N(0,I)$, the generated results sometimes show a significant color loss in the foreground object and instability in the background. To address this issue, we convert the first frame image into a latent embedding $\varepsilon_s$ and regard it as shared noise. Then, we sample a noise sequence $[\varepsilon_{2},\cdots,\varepsilon_{n}]$ from the same distribution as the base noise of samples. The shared noise is then added to the subsequent frames at a certain ratio: 
\begin{equation}  
	\varepsilon_i=\alpha\varepsilon_s+(1-\alpha)\varepsilon_i
\end{equation}
where $\alpha$ is a balance parameter. According to our experiments, setting $\alpha$ to 0.2 yields the best results. Adding the first frame's features to the other frames effectively prevents the loss of certain features during video generation and ensures continuity between frames. The complete algorithm is outlined in Alg \ref{alg:algorithm}.

\begin{algorithm}[!t]
\caption{Mask-Guided Video Generation}
\label{alg:algorithm}
\textbf{Input}: Mask $M=\{m_1,m_2,...,m_n\}$ ,Text prompt $c_p$ \\
\textbf{Parameter}: $T$\\
\textbf{Output}: $V$:generated video 
\begin{algorithmic}[1] 
\STATE $V_1=ControlNet(\varepsilon_1,c_p,m_1)$
\STATE $\varepsilon_{s}=E(V_{1})$
\FOR{$i = 2$ to $T$}
    \STATE $\epsilon_i = \alpha \epsilon_s + (1 - \alpha) \epsilon_i$
\ENDFOR
\FOR{$t = 2$ to $T$}
    \STATE $v_t = \text{MaskVideo}(c_p, M, \epsilon_t, \epsilon_s)$
    \STATE $v.\text{append}(v_t)$
\ENDFOR
\STATE $V = D(v)$
\STATE \textbf{return} $V$
\end{algorithmic}
\end{algorithm}

\section{Experiment}

\subsection{Implementation Details}
In our experiments, we train on only 5–8 videos with the same motion pattern, utilizing the SDv1-4 \cite{rombach2022high} weights to facilitate subsequent frame prediction. The model is trained for 15,000 epochs. During the inference stage, we employ ControlNet \cite{zhang2023adding} to generate the first frame, and produce the motion video by feeding in a mask sequence. We adjust only the parameters of the newly added layers, as well as those in the self-attention and cross-attention layers. The learning rate is set to $3.0\times10^{-5}$. All experiments are conducted on a single NVIDIA RTX 4060Ti, requiring approximately 15 GB of memory for training and around 12 GB for inference.

\subsection{Qualitative Evaluation}

\begin{figure}[!t]
    \includegraphics[width=\columnwidth]{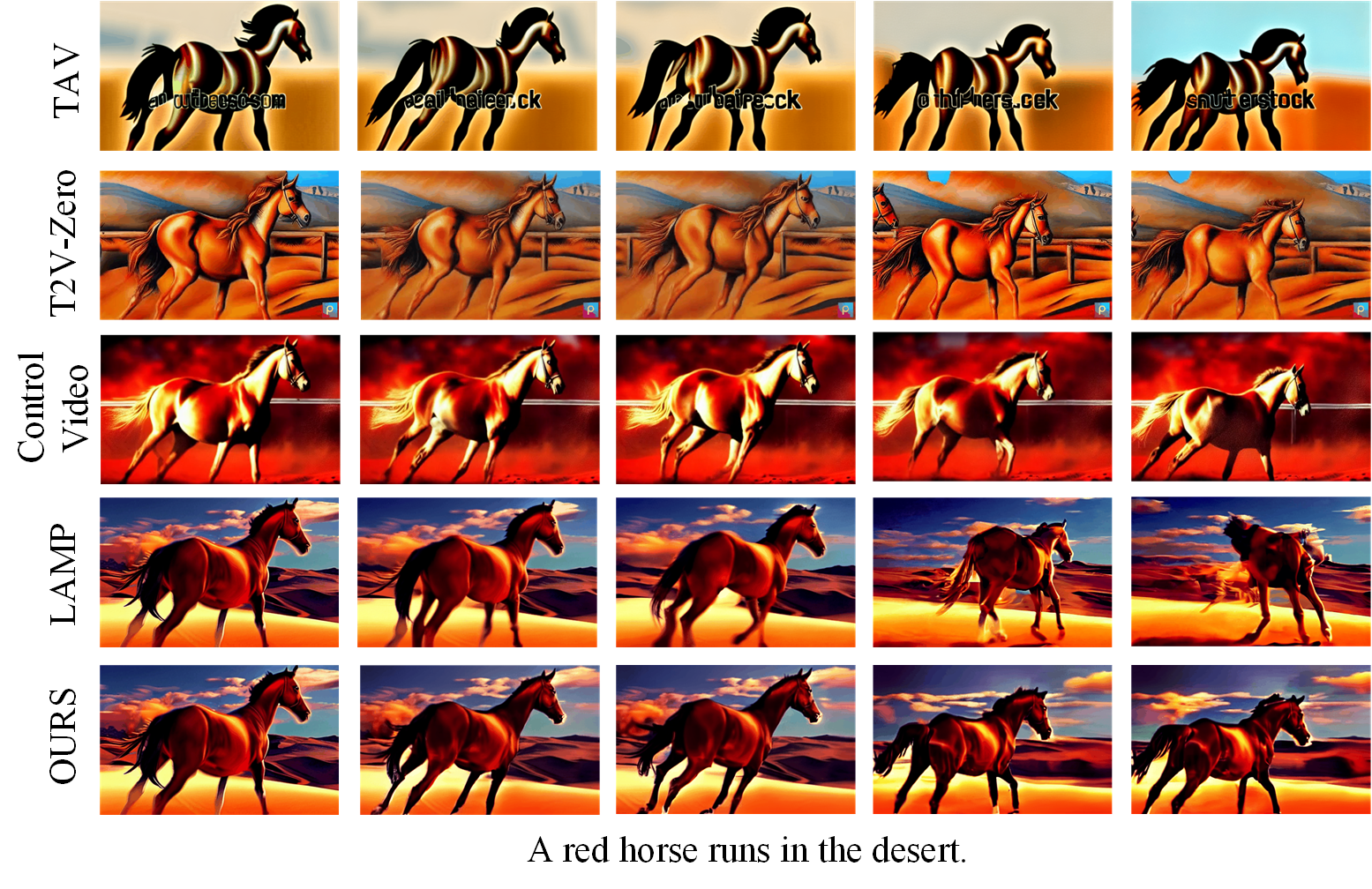}  
    \caption{Comparison between our method and baselines.}
\label{fig6}
\end{figure}
\begin{table}[!t]
\renewcommand{\arraystretch}{1.3}
\caption{Quantitative comparisons with the evaluated T2V methods}
\label{table}
\centering
\begin{tabular}{l|cc}
\hline
\textbf{Method} & \textbf{Consistency} & \textbf{Alignment} \\
\hline
Tune-A-Video & 95.4 & 30.9  \\
Text2Video-Zero & 97.0  & 32.6 \\
ControlVideo & 97.1  & 32.1 \\
LAMP & 95.1  & 32.3  \\
\textbf{Ours} & \textbf{97.4}$\uparrow$  & \textbf{33.0}$\uparrow$  \\
\hline
\end{tabular}
\end{table}
We train our model on three types of data: animal motion, multi-object motion, and rigid body motion, as shown in Fig.\ref{fig1}. We compare our method with several publicly available baselines: Tune-A-Video \cite{wu2023tune}, Text2Video-Zero \cite{khachatryan2023text2video}, ControlVideo\cite{zhang2024controlvideo}, LAMP\cite{wu2024lamp}. As shown in Fig.\ref{fig6}, Tune-A-Video \cite{wu2023tune} exhibits excessive dependence on original video content, resulting in limited generation quality. Text2Video-Zero \cite{khachatryan2023text2video} and ControlVideo \cite{zhang2024controlvideo} 
demonstrate high sensitivity to the input video, with their generated outputs frequently being influenced by the background of the provided video. For example, the appearance of a fence in the background in Fig \ref{fig6} is clearly unreasonable. LAMP \cite{wu2024lamp} exhibits limitations in capturing foreground in the first frame, leading to sudden changes and disappearance of foreground objects. In contrast, our method accurately captures the motion trajectory of foreground text and effectively distinguishes between foreground and background. Since our input motion sequences are generated through drawing or extraction without background information, our method can consistently generate video content free from background interference.
%

\subsection{Quantitative Evaluation}
 We evaluate our model in terms of text alignment and frame consistency. Text alignment is measured using the CLIP model to compute embedding similarity between generated video frames and their text descriptions \cite{radford2021learning}. Frame consistency quantifies continuity by comparing embedding similarities between adjacent frames using the CLIP model. In our experiments,  we used different motion sequences to generate 26 distinct results and calculate the average results. As shown in Table \ref{table}, our method outperforms other baselines in both text alignment and frame consistency metrics.



\begin{figure}[!t]
    \includegraphics[width=\columnwidth]{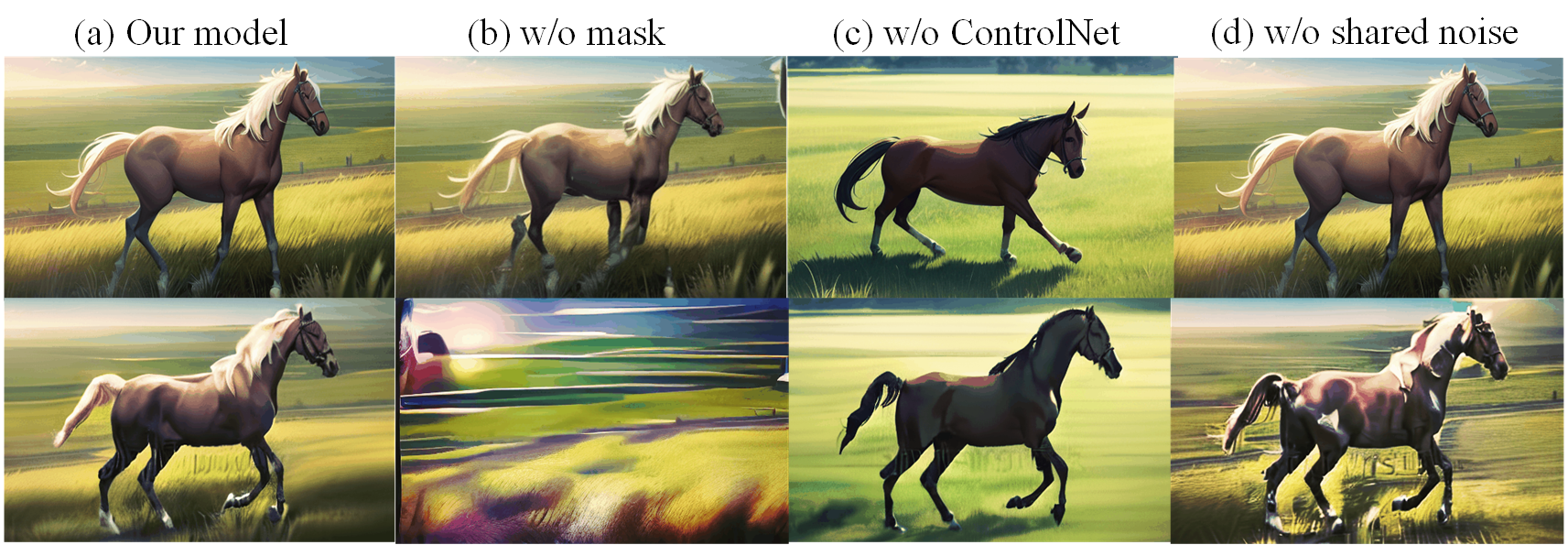}  
    \caption{\textbf{Ablation results.} The given prompt is ‘A horse runs on the grass’. }

\label{fig7}
\end{figure}

\begin{table}[!t]
\renewcommand{\arraystretch}{1.3}
\caption{Ablation Studies on Different Components}
\label{table2}
\centering
\begin{tabular}{l|ccc}
\hline
 type & IoU  & Text  & Frame  \\
\hline
w/o mask & 0.184 & 0.261 & 0.867 \\
w/o controlnet & 0.864 & 0.299 & 0.962 \\
w/o sharing & 0.908 & 0.302 & 0.966 \\
Full model & \textbf{0.917}$\uparrow$ & \textbf{0.310}$\uparrow$ & \textbf{0.975}$\uparrow$ \\
\hline
\end{tabular}

\end{table}


\subsection{Ablation Study}

We remove key components to observe their impact on video quality. As shown in Fig.\ref{fig7}, removing the motion sequence mask causes the disappearance of target foreground objects. Without ControlNet\cite{zhang2023adding} in first-frame generation, we observe significant misalignment between the initial frame and the mask, which severely impacts the overall video quality. Additionally, disabling the first-frame sharing strategy results in noticeable color inconsistencies in foreground objects. For quantitative evaluation, we measure the frame consistency and text alignment of the foreground region, as well as the IoU precision between the generated video's foreground and the input mask. Table \ref{table2} presents the detailed quantitative results.

\section{Conclusion}
In this letter, we present a mask-guided video generation approach that requires only limited data and a single GPU for training. By incorporating foreground masks and enhanced regional attention, our method improves video quality and motion control. For future work, we plan to explore simpler mask sequence generation methods to make the system more accessible for video creation and editing applications.

\newpage
\balance  
\bibliographystyle{IEEEtran}
\bibliography{IEEE_conference}

\end{document}


\title{Supplementary Materials for\\\textit{``Resource-Efficient Motion Control for Video Generation via Dynamic Mask Guidance"}}

\author{Sicong Feng, Jielong Yang, and Li Peng}
\maketitle

\begin{figure}[!t]
    \includegraphics[width=\columnwidth]{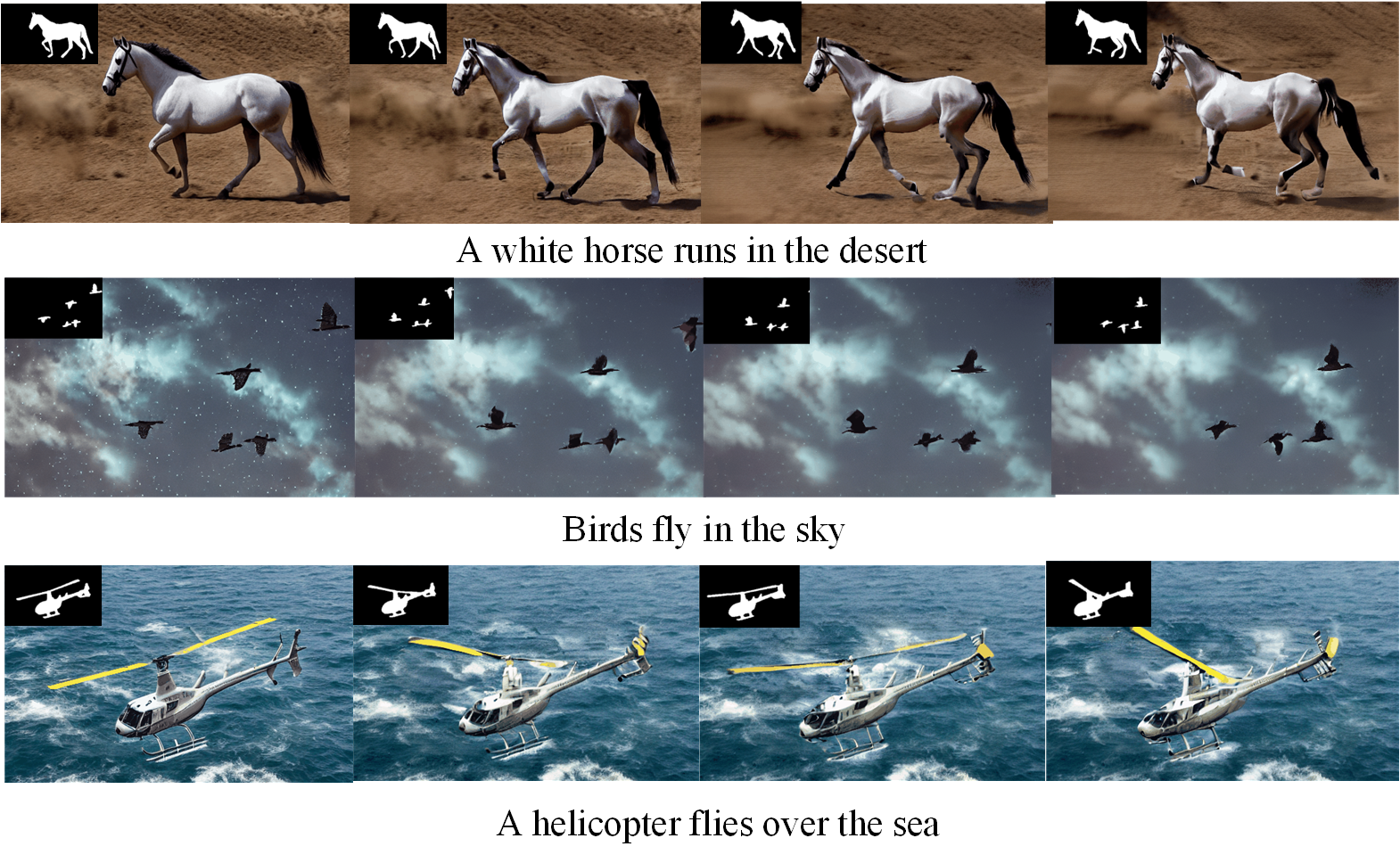}
    \caption{The resulting images generated from the drawing.}
    \label{fig2}
\end{figure}

\section{interactive editing}

Our proposed mask-guided video generation framework not only
achieves high-quality video synthesis but, more significantly,
introduces a simple yet effective interactive editing mechanism.
The binary nature of our mask representation, consisting of
only black and white values, enables users to create desired
motion sketches through intuitive drawing operations, as
illustrated in Fig. 1.

Traditional text-to-video generation methods primarily rely on
textual prompts for content control, often lacking precision
and intuitiveness. In contrast, our framework, through
interactive mask editing, enables precise control over multiple
key attributes in the generated videos. As demonstrated in
Fig. 2, users can control the size variations of foreground
objects through simple scaling operations. Furthermore, by
adjusting the direction and spacing of mask sequences, users
can directly manipulate the direction and velocity of motion,
providing significantly more precise and predictable control
compared to purely text-based descriptions. Most notably, our
method enables flexible control over foreground object quantity.
Users can easily add multiple objects and independently control
their trajectories through mask manipulation. This interactive
approach enhances practical video generation applications by
combining intuitive controls with high-quality output.

\begin{figure}[!t]
    \includegraphics[width=\columnwidth]{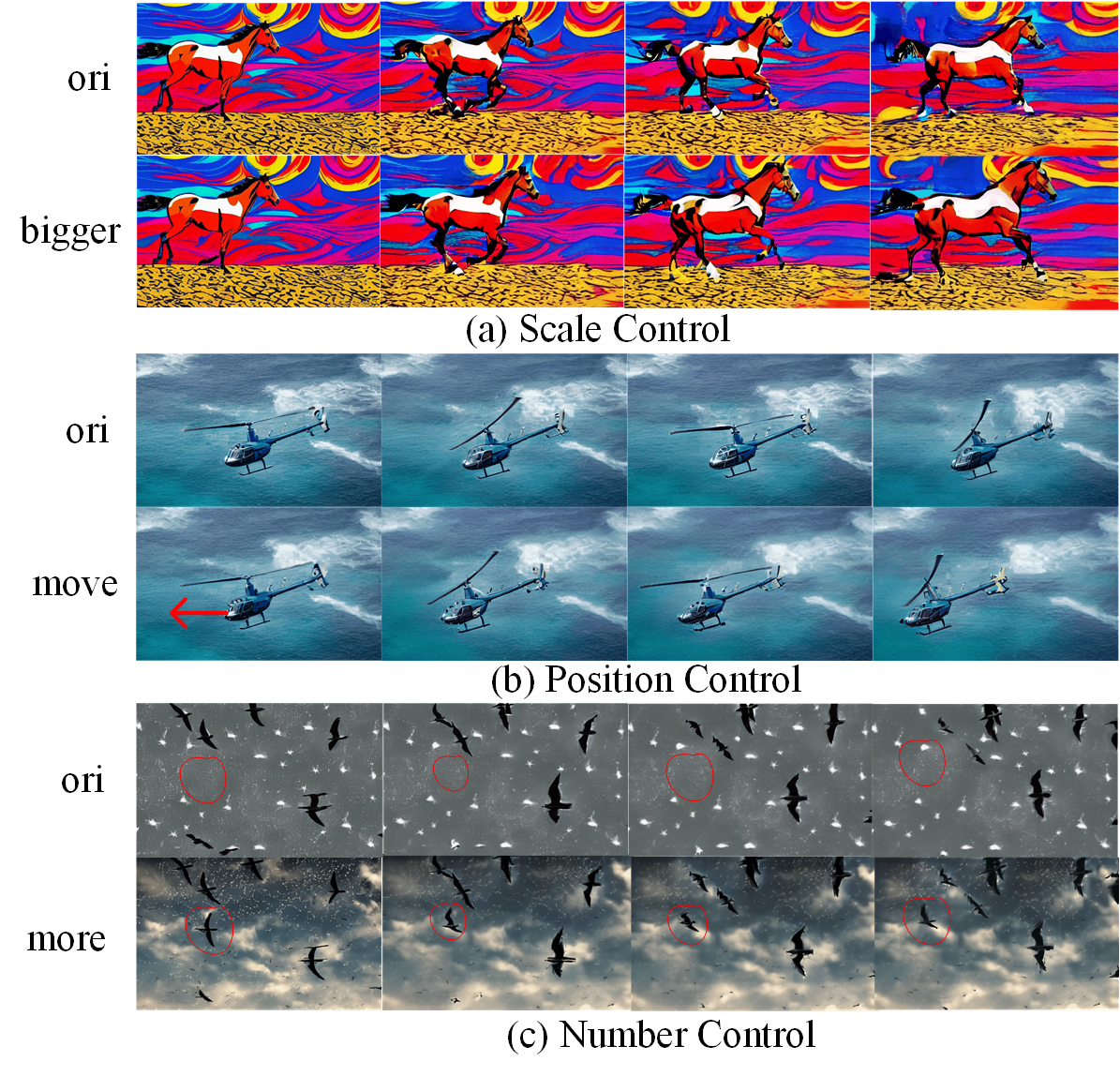}
    \caption{Interactive mask editing for precise video Generation.}
    \label{fig3}
\end{figure}



\section{user study}

We conduct a user evaluation survey to compare the performance of our method with other publicly available generation methods. Specifically, we create a questionnaire using 22 video samples, providing each evaluator with a set of text prompts and the corresponding generated results. We ask them to select the better generated videos based on two criteria: video quality and the alignment between the prompts and the generated videos. Ultimately, we receive 33 completed questionnaires. As shown in Table \ref{table}, evaluators prefer our generated videos in both aspects. In contrast, Tune-A-Video, which only uses DDIM inversion for structural guidance, fails to produce consistent and high-quality videos, while the videos generated by Text2Video-Zero also exhibit lower quality.


\begin{table}[!t]
\renewcommand{\arraystretch}{1.3}

\caption{User preference comparisons with the evaluated T2V methods}
\centering
\begin{tabular}{l|c|c}
\hline
Method & Frame Consistency & Textual Faithfulness \\
& (User Preference) & (User Preference) \\
\hline
Tune-A-Video & 11.0 & 12.4 \\
Text2Video-Zero & 9.4 & 12.4 \\
LAMP & 24.5 & 22.2 \\
OURS & \textbf{55.1} & \textbf{53.0} \\
\hline
\end{tabular}
\label{table}
\end{table}